\documentclass{article}

\usepackage{arxiv}

\usepackage[utf8]{inputenc} 
\usepackage[T1]{fontenc}    
\usepackage{hyperref}       
\usepackage{url}            
\usepackage{booktabs}       
\usepackage{amsfonts}       
\usepackage{nicefrac}       
\usepackage{microtype}      
\usepackage{lipsum}
\usepackage{graphicx}
\graphicspath{ {./images/} }
\usepackage{cite}
\usepackage{amsmath,amssymb,amsfonts}
\usepackage{algorithm}
\usepackage{algorithmic}
\usepackage{graphicx}
\usepackage{textcomp}
\usepackage{xcolor}

\title{Autonomous Generation of Sub-goals for Lifelong Learning in Robots}

\author{
Emanuel Fallas Hernández\\
GII, CITIC research center \\
Universidade da Coruña\\
A Coruña, Spain \\
\texttt{emanuel.fallas@udc.es}\\
   \And
Sergio Martínez Alonso\\
GII, CITIC research center \\
Universidade da Coruña\\
A Coruña, Spain \\
\texttt{sergio.martinez3@udc.es}\\
  \And
Alejandro Romero\\
GII, CITIC research center \\
Universidade da Coruña\\
A Coruña, Spain \\
\texttt{alejandro.romero.montero@udc.es}\\
  \And
Jose A. Becerra Permuy\\
GII, CITIC research center \\
Universidade da Coruña\\
A Coruña, Spain \\
\texttt{jose.antonio.becerra.permuy@udc.es}\\
  \And
Richard J. Duro\\
GII, CITIC research center \\
Universidade da Coruña\\
A Coruña, Spain \\
\texttt{richard.duro@udc.es}\\
}

\begin{document}
\maketitle
\begin{abstract}
One of the challenges of open-ended learning in robots is the need to autonomously discover goals and learn skills to achieve them. However, when in lifelong learning settings, it is always desirable to generate sub-goals with their associated skills, without relying on explicit reward, as steppingstones to a goal. This allows sub-goals and skills to be reused to facilitate achieving other goals. This work proposes a two-pronged approach for sub-goal generation to address this challenge: a top-down approach, where sub-goals are hierarchically derived from general goals using intrinsic motivations to discover them, and a bottom-up approach, where sub-goal chains emerge from making latent relationships between goals and perceptual classes that were previously learned in different domains explicit. These methods help the robot to autonomously generate and chain sub-goals as a way to achieve more general goals. Additionally, they create more abstract representations of goals, helping to reduce sub-goal duplication and make the learning of skills more efficient. Implemented within an existing cognitive architecture for lifelong open-ended learning and tested with a real robot, our approach enhances the robot's ability to discover and achieve goals, generate sub-goals in an efficient manner, generalize learned skills, and operate in dynamic and unknown environments without explicit intermediate rewards.
\end{abstract}


\section{Introduction} 

\label{sec:intro}
Cognitive architectures serve as the foundation for intelligent robotic systems \cite{kotseruba202040}. They enable the integration of perception, reasoning, and learning to facilitate goal-oriented behavior in complex, dynamic environments. From their beginnings as static, rule-based architectures \cite{lebiere1993connectionist, varma20064caps, laird1987soar} to current systems integrating machine learning algorithms and neural networks for continuous adaptation and learning \cite{kotseruba202040, becerra2021motivational, romero2022autonomous, weng2004developmental}, these architectures provide the computational structures and algorithms necessary for robots to process sensory data, form abstract representations, and make decisions that guide their actions. 

Central to these architectures is the concept of perceptual classes \cite{feldman1992constructing}, which are subsets of the robot's state space that represent clusters of sensory data or environmental features that are perceptually or functionally similar. Perceptual classes are fundamental in bridging raw sensor data and higher-level reasoning, as they allow robots to group related sensory states into actionable categories.

Perceptual classes can be formally defined as subsets of the robot's state space. Let \( S \) denote the robot's state space, which encompasses all possible configurations of the robot and its environment as perceived through its sensors. A perceptual class \( \hat{s} \) is defined as:
\[
\hat{s} \subseteq S
\]
where each \( s \in \hat{s} \) corresponds to a sensory state that the robot's perception system classifies as belonging to \( \hat{s} \). The set of all perceptual classes is denoted \( \hat{S} \), and the mapping from the robot's sensory state \( s \in S \) to one or more perceptual classes \( \hat{s} \in \hat{S} \) is defined by the perception function:
\[
P: S \to \mathcal{P}(\hat{S})
\]
where \( \mathcal{P}(\hat{S}) \) is the power set of \( \hat{S} \). This function enables the robot to map raw sensory inputs to perceptual classes, forming the basis for reasoning and decision-making.

Goals in cognitive architectures are another fundamental concept. A goal represents a desired state or set of states in the robot's state space \( S \). In this sense, they represent a subset of perceptual classes, those containing the states that at some point the system needs to reach. Formally, a goal \( g \) can be defined as:
\[
g \subseteq S
\]
where \( g \) represents a subset of the state space that the robot seeks to achieve or maintain. Goals often interact with other perceptual classes, as the robot's ability to recognize relevant perceptual classes is essential for determining the actions required to transition toward the goal state. For example, if the robot's goal is to navigate to a specific location, perceptual classes such as ``path'', ``obstacle'', and ``landmark'' are critical for planning and execution.

The ability to autonomously generate perceptual classes is a crucial feature of advanced cognitive architectures. It is what allows them to transform continuous perceptual experiences into representations that permit discrete categorical decision-making, bridging raw sensory data and symbol-like representations \cite{coradeschi2013short}. Rather than relying on predefined categories, robots operating in real-world environments must dynamically create, refine, and adapt perceptual classes through continuous interaction with their surroundings in order to achieve an appropriate grounding of their knowledge \cite{steels2008symbol, cangelosi2010grounding}. This process allows robots to accommodate novel situations and domains that were not anticipated during their design or training.

Autonomously generated perceptual classes are particularly valuable in scenarios where the robot's environment changes over time or includes elements that cannot be exhaustively pre-specified. For instance, a household robot might initially classify objects broadly into ``furniture'' or ``tools'', but through repeated interactions, it could autonomously refine these classes into more specific subcategories, such as ``chairs'', ``tables'', ``screwdrivers'', and ``wrenches'', based on observed features and functional distinctions.

The process of autonomously generating perceptual classes is closely linked to lifelong learning \cite{Thrun1998lifelong}, wherein robots continuously acquire and refine knowledge throughout their operational lifespan. Unlike conventional machine learning paradigms, which rely on static datasets and predefined curricula, lifelong learning reflects the dynamic and incremental nature of real-world robotic applications, endowing them with the ability to engage in open-ended learning \cite{sigaud2023definition}. This leads to the concept of lifelong open-ended learning autonomy (LOLA) \cite{romero2023perspective} as a goal to strive for.

In this context, the robot's learning trajectory is not predetermined but is constructed in real-time, driven by the situations it encounters and the goals it seeks to achieve. This dynamic learning process requires the robot to identify and form new perceptual classes and discover goals on-the-fly. Formally, let \( \{t_1, t_2, \ldots, t_n\} \) represent discrete time steps. At each time step \( t_i \), the robot observes a subset of the state space \( S_i \subseteq S \) and generates or refines perceptual classes \( \hat{S}_i \subseteq \hat{S} \) and goals \( G_i \subseteq G \). The cumulative learning process over time is represented as:
\[
\bigcup_{i=1}^n \hat{S}_i \to \hat{S}, \quad \bigcup_{i=1}^n G_i \to G
\]

Despite the asynchronous and dynamic nature of lifelong learning, it is desirable to organize the robot's accumulated knowledge into a coherent structure. One effective approach is to represent perceptual classes and goals as nodes in graphs \cite{soundararajan2003depth} \cite{van2008perceptual}, with edges capturing known relationships between them. Formally, let \( G = (V, E) \) represent a directed graph, where:
\begin{itemize}
    \item \( V \) is the set of nodes, representing perceptual classes \( \hat{s} \in \hat{S} \) and goals \( g \in G \).
    \item \( E \) is the set of edges, where \( (v_1, v_2) \in E \) denotes a relationship between two nodes \( v_1, v_2 \in V \).
\end{itemize}

An additional challenge lies in discovering relationships between initially disconnected subgraphs. By analyzing the state space regions associated with different perceptual classes or goals, the robot can identify overlaps or transitions that suggest latent connections. This process can be formalized by identifying common elements in the state space:
\[
\exists s \in S : s \in S_{\hat{s}_1} \cap S_{\hat{s}_2}
\]

Such connections allow the robot to integrate disparate knowledge into a unified framework, supporting generalization and cross-domain reasoning.

Perceptual classes and goals, defined as regions of a robot's state space, are fundamental to cognitive architectures that enable autonomous behavior. The dynamic generation of perceptual classes and the asynchronous acquisition of goals through lifelong learning allow robots to adapt to new environments and tasks. Organizing this knowledge through graph-based representations provides a structured means to manage complexity, discover latent relationships, and support flexible decision-making. These capabilities are critical for developing intelligent robotic systems capable of operating in complex, unstructured environments. However, discovering latent relationships when considering LOLA robots is not evident. The robot is constantly learning, leading to perceptual classes and goals that are not completely defined. Additionally, long goal chains are hard to establish.

In this paper we propose a two pronged approach to the generation and discovery of these latent relationships among goal and non-goal perceptual classes: a top-down approach, where sub-goals are hierarchically derived from general goals based on intrinsic motivations, and a bottom-up approach, where sub-goal chains emerge from goals and non-goal perceptual classes that were previously learned in different domains. These methods help the robot autonomously generate and chain sub-goals as a way to achieve more general goals. 

The rest of the paper is structured as follows: Section \ref{sec:e-MDB} provides an overview of the cognitive architecture where our approach to autonomous sub-goal generation is applied and evaluated. Section \ref{sec:sub-goal_generation} explains the specific methodology used for generating sub-goals, detailing the two strategies implemented: a top-down approach and a bottom-up approach. Section \ref{sec:experiment} presents a robotic experiment to demonstrate this approach, along with a discussion of the most significant results. Finally, Section \ref{sec:conclusion} offers some conclusions and explores the implications of incorporating the proposed mechanism into cognitive architectures.

\section{e-MDB cognitive architecture}
\label{sec:e-MDB}

The cognitive architecture we use to test our approach is the epistemic Multilevel Darwinist Brain (e-MDB) \cite{becerra2021motivational}. This architecture was designed to address the problem of lifelong open-ended learning autonomy (LOLA) in robotic systems, and it was chosen because of its ability to autonomously acquire and relate different knowledge nodes, including both perceptual classes and goals.

To achieve this, the e-MDB includes a Motivational System \cite{romero2020motivation}, which manages the motivations of the robot, facilitating both the discovery and activation of goals as well as the autonomous generation of perceptual classes. It is composed of domain-independent needs and drives \cite{hawes2011survey} that gradually align with the specific goals identified by the robot in different domains. This enables the robot to operate in open-ended learning environments without requiring prior knowledge of the goals it needs to achieve.

The knowledge discovered and generated by the Motivational System is then refined through the Learning System, which is responsible for acquiring the various elements of knowledge necessary for the robot to function across multiple domains. For instance, it enables the robot to learn skills to achieve the discovered goals, construct world models that represent the behavior of its environment, and continuously refine perceptual classes through interaction with the environment. The skills can be learned in the form of utility models \cite{romero2019simplifying} or policies \cite{romero2023using}.

\begin{figure}
    \centering
    \includegraphics[width=0.52\textwidth, trim={5cm 5cm 5cm 5cm},clip]{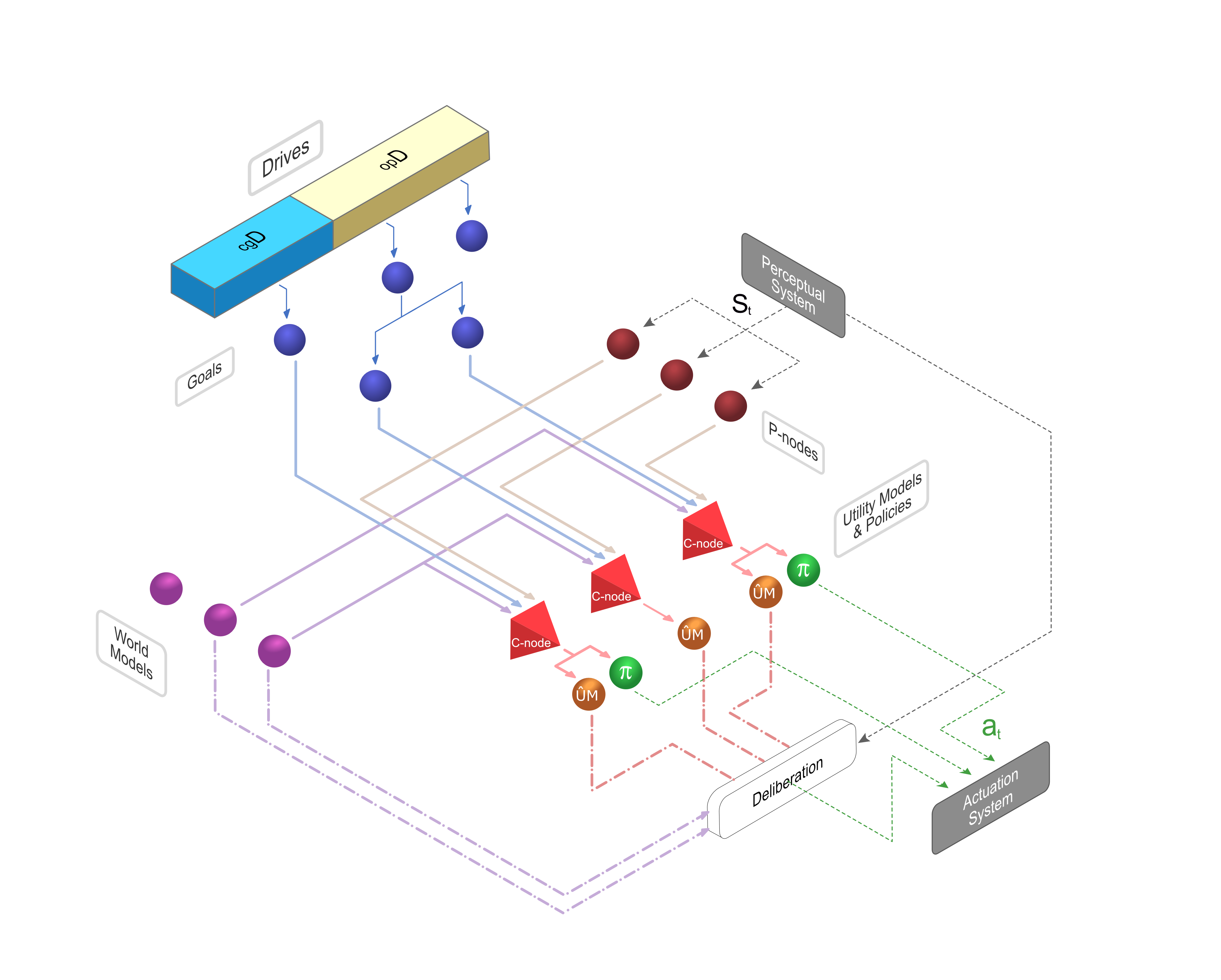}
   \caption{Memory structure of the e-MDB cognitive architecture.}
    \label{Fig:memory_schematic}
\end{figure}

All the knowledge is organized in a Memory System \cite{duro2019perceptual}, which acts as the central element of the architecture. This system is responsible for decision-making regarding which actions to execute. Its structure is shown in Fig. \ref{Fig:memory_schematic} where the relationships between the various knowledge nodes are represented as a directed graph, as explained in Section \ref{sec:intro}. The specific knowledge nodes include:

\begin{itemize}
    \item Drives ($D$): domain-independent indicators of the system's deviation from desired states or needs.
    \item Goals ($G$): Target states in the perceptual space that, when achieved, reduce the value of certain drives ($D$). 
    \item World Models ($WM$): Models that characterize the behaviour of the domain, predicting the next state ($S_{t+1}$) based on current state ($S_t$) and the applied action ($a_t$).
    \item Utility Models ($\hat{U}M$): Functions associated with a goal ($G$) that estimate the expected utility of perceptual states ($S_t$) in a given domain.
    \item Policies ($\pi$): Reactive representations that provide an action ($a_t$) for every perceptual state ($S_t$) within a specific perceptual class, related to achieving a specific goal ($G$).
    \item P-nodes ($P_n$): Represent perceptual classes, grouping perceptions that lead to the same next perceptual state ($S_{t+1}$), or perceptual class, when the same action ($a_t$) is applied.
   \item C-nodes ($C_n$): Contextual nodes that link P-nodes, goals, and the utility models or policies needed to transition between initial perceptual states (represented by P-nodes) and target perceptual states (represented by goals) within a specific domain (represented by a world model). 
\end{itemize}

The operation of the architecture is centered around C-nodes, represented as red pyramids in Fig. \ref{Fig:memory_schematic}. These nodes, functioning as product-type units and are activated when their input nodes ($WM$, $G$, and $P_n$) are activated, propagating their activation to the nodes connected to their output ($\hat{U}M$ and/or $\pi$). That is, in a given domain (represented by its $WM$), when the robot perceives a state corresponding to a certain perceptual class ($P_n$) and a specific goal ($G$) is active, the C-node triggers the associated policy ($\pi$) or utility model ($\hat{U}M$). Thus, if multiple policies or UMs are activated, the robot's behavior will be guided by the policy or utility model with the highest level of activation, thereby determining its actions.

As shown in Fig. \ref{Fig:memory_schematic}, both P-nodes (dark red circles) and goals (blue circles) are key elements in the architecture's operation. 
The correct modeling, linking, and activation of these nodes will determine the robot’s ability to perform appropriate actions under specific perceptual conditions and achieve its goals. Moreover, if domains or tasks change, it is crucial that both perceptual classes and goals can be reused and updated.

In the following section we present our proposed approach for the autonomous and online generation of sub-goals, where the concepts of goals and perceptual classes will play a central role.

\section{Approach to Sub-Goal Generation}
\label{sec:sub-goal_generation}

In order to provide appropriate methods to sub-goal generation for lifelong learning, generalization and reusability must be considered. Generalization is required so that identified sub-goals include all possible states leading to the upstream goal, and to avoid excessive fragmentation of goals. On the other hand, reusability is required so that both goals and sub-goals learned in different domains can be leveraged in new domains when the knowledge they contain is useful. 

In our approach we leverage the use of perceptual classes within the e-MDB architecture to create new goals or to link goals that were not previously related. Perceptual classes represent a subset of the robot's state space in which specific conditions have happened. The interpretation of the points within a perceptual class depends on the type of node to which the perceptual class belongs. In the case of P-Nodes, the states contained have led to a goal when performing a specific action in a specific domain, that is, it defines preconditions to achieve a goal. The perceptual classes that represent goals, include all states which have led to obtaining a reward. 

It is important to note that perceptual classes are built during the lifetime of the robot and they need to be learned progressively. Methods for this process have been described before \cite{martinez2024Perceptual}. When presented with a new domains, the robot will have incomplete representations, for this reason, learning metrics have been devised to identify when the knowledge contained in a perceptual class has been stabilized. A confidence value for the validity of a perceptual class $C$ can be obtained using the history of previous classifications as shown in (\ref{eq:confidence}) where $P_{t-i}$ is the success of the classification (1 or 0) and $N$ is number of previous classifications considered.

\begin{equation} \label{eq:confidence}
    C=\frac{\sum\limits_{i=1}^{N} P_{t-i}}{N}
\end{equation}

\subsection{Top-Down Approach}

This approach allows creating chains of sub-goals starting from the knowledge obtained about how to reach a final goal. This is done by creating goals that aim to reach the states that correspond to previously learned P-Nodes (the starting states to reach the fi goal), so that the final goal can be achieved. Once these sub-goals are created, the robot will learn ways to reach them, generating more P-Nodes (starting states for reaching these sub-goals) that once learned can be used to create more sub-goals. This mechanism is based on the effectance cognitive drive \cite{romero2020motivation}, where reaching certain perceptual points generates an effect in the nodes existing in the architecture, in this case, P-Nodes.

\begin{figure}[ht]
  \centering
  \includegraphics[width=0.9\columnwidth]{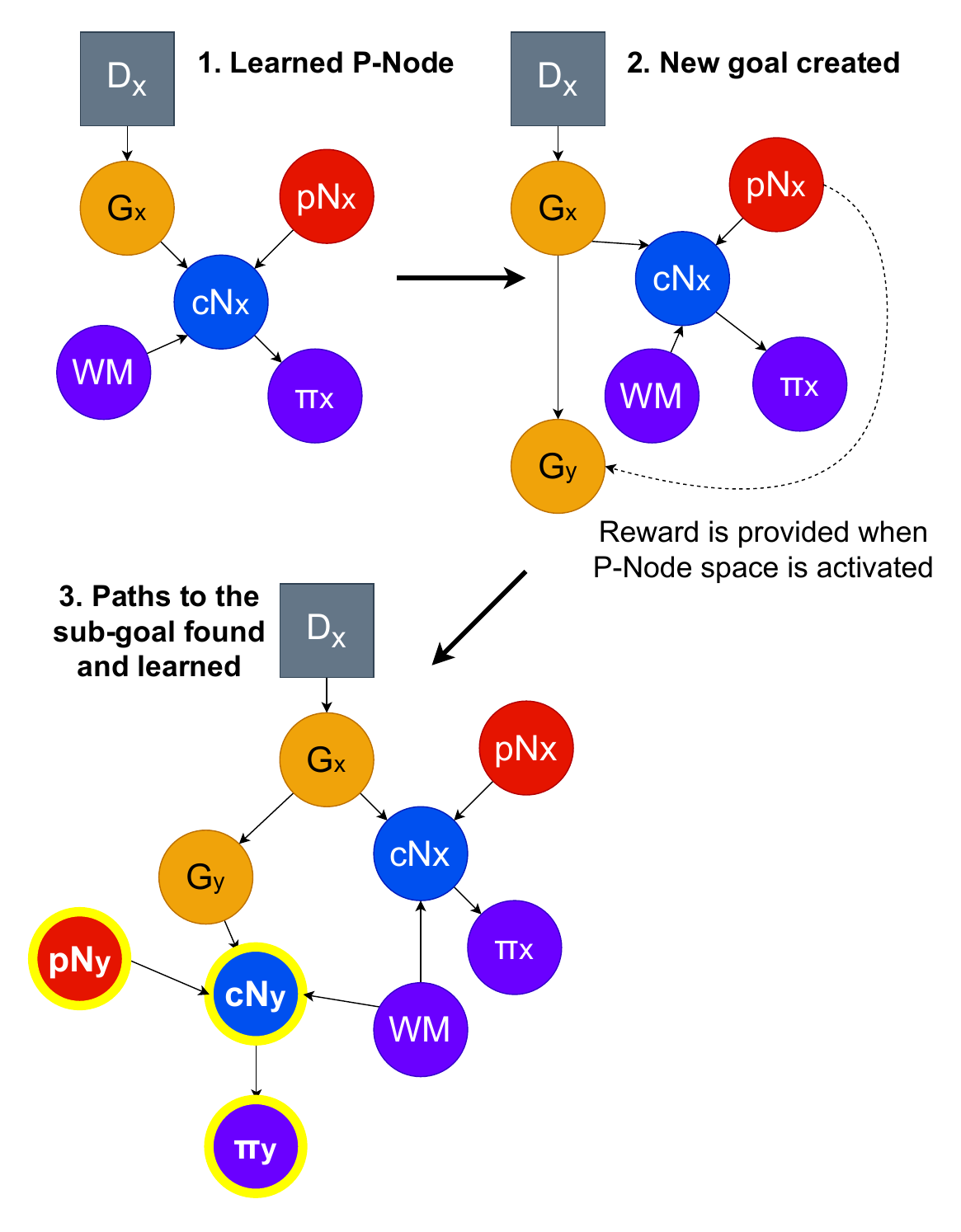}
  \caption{Top-down sub-goal generation.}
  \label{fig:topdown_subgoals}
\end{figure}

Operationally, this approach is implemented as a cognitive drive that monitors the confidence of existing P-Nodes in the architecture, once any P-Node has exceeded a threshold value, the drive will become active, as shown in (\ref{eq:eff_internal_drive}). In turn, this will activate a policy that creates a goal that will provide reward each time the related P-Node is activated. An overview of the top-down sub-goal generation is presented in Fig. \ref{fig:topdown_subgoals}.

\begin{equation} \label{eq:eff_internal_drive}
    D_{eff\_int} = 
    \begin{cases} 
    1 & \text{if any P-Node is learned,} \\
    0 & \text{otherwise.}
    \end{cases}
\end{equation}

\subsection{Bottom-Up Approach}
This approach uses the latent knowledge accumulated in the perceptual classes of P-Nodes and goals to search for new paths to achieve goals by reusing experiences learned when in different domains or while pursuing a different goal. For this, the existing relationships between the spaces of the P-Nodes and goals are analyzed. Consider a P-Node $P_i$ which is associated to goal $G_i$ through its C-Node and a goal $G_j$, the following scenarios are possible:

\begin{enumerate}
    \item P-Node i contains Goal j ($P_i \subseteq G_j$): Indicates that by reaching the goal, the P-Node will necessarily be activated. This relationship indicates that $G_j$ is a sub-goal of $G_i$ since reaching it will lead to the conditions where the upstream goal can be achieved.
    \item Goal j contains P-Node i ($G_j \subseteq P_i$): This indicates that activating the P-Node necessarily means that the goal has been reached. This condition does not imply that reaching $G_j$ will activate the P-Node in every case, but it still provides some probability of reaching it. For this reason, this case will also be considered to imply that $G_j$ is a sub-goal of $G_i$. 
    \item Overlapping but not contained Goal and P-Node: This condition will not be considered as it does not provide hierarchical information about the goals.
    \item Disjoint Goal and P-Node: This condition will not be considered as it does not provide hierarchical information about the goals.
\end{enumerate}

An alternative way of understanding this approach is as a prospection process applied to the high-level information contained in the spaces of the P-Nodes and goals. In this manner, possible states to be reached (goals) are evaluated for their utility to reach a different goal (activation of a P-Node). These paths to goals are reinforced by cascading the activation from the upstream goals to the downstream goals. 

\begin{figure}[ht]
  \centering
  \includegraphics[width=0.7\columnwidth]{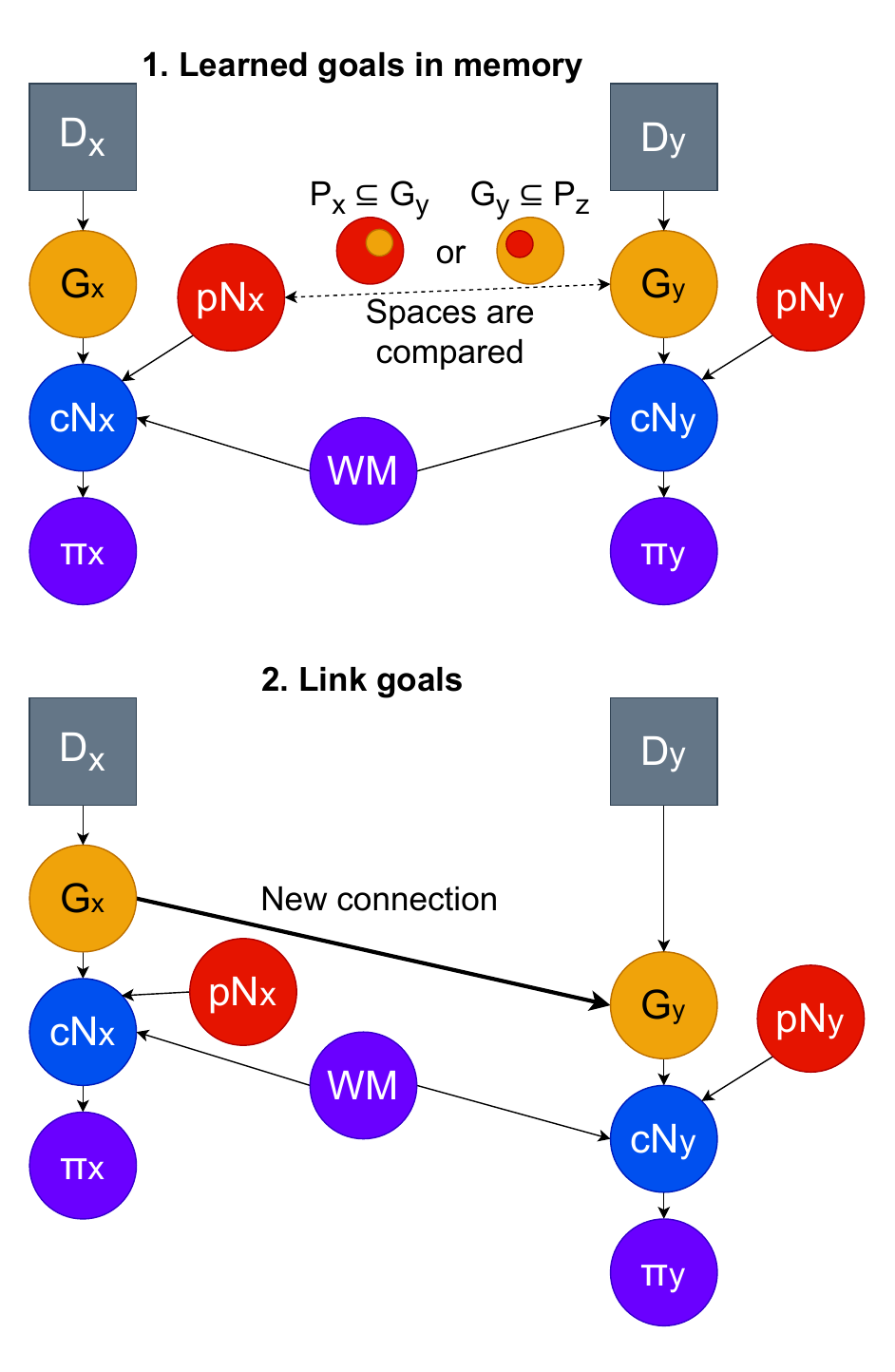}
  \caption{Bottom-up sub-goal generation. $G_y$ is identified as sub-goal of $G_x$.}
  \label{fig:bottomUp_subgoals}
\end{figure}

The implementation of this method follows the same pattern as the top-down method, where a cognitive drive searches every learned goal and P-Node for containment. As shown in (\ref{eq:prospection}), the drive will become active when a match is found. This will activate a policy which will link the upstream goal to the downstream goal, so that the activation is cascaded. The overall process for bottom-up sub-goal generation is shown in Fig. \ref{fig:bottomUp_subgoals}.

\begin{equation} \label{eq:prospection}
    D_{prospection} = 
    \begin{cases} 
    1 & \text{if a sub-goal is found,} \\
    0 & \text{otherwise.}
    \end{cases}
\end{equation}

It is important to constrain the generation of sub-goals so that no undesirable effects are produced. In order to avoid these pitfalls, mechanisms were devised to prevent these when linking sub-goals. To avoid self-feedback of goal activation, when linking a new goal, the whole chain is analyzed for loops, in case they exist, the link is not performed. Secondly, to prevent deprioritization of the top-level goals, when cascading activation from a goal to a sub-goal, an attenuation term is included so that the sub-goals are always less active than the final goal, this would make the robot prioritize achieving the final goal if steps can be skipped. 

\section{Robotic Experiment}
\label{sec:experiment}

In this section, we present an experiment where a task consisting of multiple steps is learned by generating space representations of the different sub-goals required by using both the top-down and the bottom-up approaches presented in section \ref{sec:sub-goal_generation}. We show how after interacting with the environment, a goal tree is developed by the architecture, which will demonstrate the capability of the system to represent the steps required for reaching a goal.

These experiments were integrated into the e-MDB architecture presented in section \ref{sec:e-MDB}. It is assumed that all the other elements of the architecture operate properly and, therefore, only the performance related to the goal generation will be analyzed. 

\subsection{Experimental Setup}
\begin{figure}[ht]
  \centering
  \includegraphics[width=1.0\columnwidth]{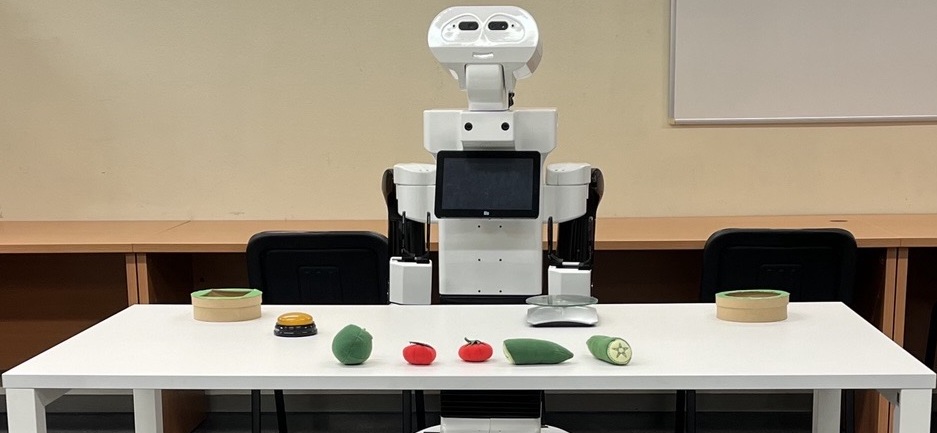}
  \caption{Experimental setup with a TIAGo robot.}
  \label{fig:experimental_setup}
\end{figure}

In this experiment the final objective is to classify fruits, as accepted or rejected, based on their weight. For that, we designed the setup shown in Fig. \ref{fig:experimental_setup}.

As can be seen, the fruits are located in a constrained area at the edge of the table, on the side furthest from the robot, called the collection area. The robot has to pick a fruit from that area and place it on the scale, which is located somewhere within another delimited area, called the weighing area, to determine whether it should be accepted or not. Finally, if the fruit is valid, the robot has to pick and leave it in the basket on its right side. If not, the fruit must be placed in the basket on its left side. There is also a button on the table that, when pressed, switches its light on/off, but does not contribute to the classification task.

In this setup, an RGB camera, located on the ceiling of the room, provides the robot with the position of the scale and the fruit closest to the robot. That position is given in polar coordinates, with angles between -1.4 and 1.4 radians and distances between 0.2 and 1.9 meters. The camera also gives us the value of the maximum dimension of the perceived fruit, which is between 0.03 and 0.1 meters; and if the light of the button is on or not.

We also know if a fruit is grasped by one of the robot's grippers and the state of the scale. When a fruit is tested, the latter is active and it has a state that represents whether a fruit is valid (1) or not (2). If a fruit has not been tested yet, it is inactive and the state is 0. Thus, the cognitive architecture works with a ten-dimensional perceptual state, in which all perceptions are normalized between 0 and 1.

To complete the task, 8 policies, described in Table \ref{tab:policies}, were designed. Using \textit{Pick fruit}, the robot can grasp a piece of fruit anywhere on the table, within its reach. When a fruit is grasped, \textit{Test fruit} can be used to place it on the scale and evaluate it. Finally, with \textit{Accept fruit} and \textit{Discard fruit}, the fruit is placed in the respective basket. Since the robot does not reach the entire setup with one arm, it sometimes has to change the piece of fruit from one gripper to another. If the maximum dimension of the any particular fruit is less than 0.085 meters, \textit{Change hands} can be used; if not, \textit{Place fruit} must be used to leave it in the middle of the table and pick it up with the other arm. \textit{Ask nicely} is used if there are no fruits in the collecting area and \textit{Press button} can be used to turn on/off the light.

In this experiment, all the perceptual classes are modeled with neural networks with 3 dense hidden layers, plus the input and output layers. All of them use a ReLU activation function, except the last one, which uses a sigmoid and determines the activation of the perceptual classes. The number of neurons per layer is 10, 128, 64, 32 and 1, from input to output. Since the labels used to train are 0 (point does not belong to the perceptual class) or 1 (it belongs), we use the binary crossentropy loss function for training, in addition to the Adam optimizer. The input dataset is limited to 400 entries to avoid problems of overfitting, behaving like a FIFO queue when that limit is reached. The batch size and the number of epochs are 50. A weighting policy is used to compensate for the unbalanced nature of the dataset. Finally, the training phase is only done when the output differs from the prediction of the network.

\begin{table}
    \centering
    \caption{Policies available}
    \label{tab:policies}
    \begin{tabular}{|p{2cm}|p{5,6cm}|}
        \hline
        \textbf{Policy} & \textbf{Description}\\
        \hline
         \textit{Pick fruit} & Use a gripper to grasp a fruit\\
         \hline
         \textit{Test fruit}& Leave the fruit in the scale to evaluate it\\
         \hline
         \textit{Accept fruit} & Leave the fruit in the accepted basket\\
         \hline
         \textit{Discard fruit} & Leave the fruit in the rejected basket\\
         \hline
         \textit{Change hands} & Move the fruit from one gripper to the other\\
         \hline
         \textit{Place fruit} & Leave the fruit in the middle of the table\\
         \hline
         \textit{Ask nicely} & Ask experimenter to provide more fruits, if none\\
         \hline
         \textit{Press button} & Press the button to turn on/off its light\\
         \hline
    \end{tabular}
\end{table}

\subsection{Compared Approaches}
In order to assess the effectiveness of our approaches for sub-goal generation, a first experiment was carried out to establish a baseline for comparison purposes. In this baseline experiment, the robot is not autonomous, as a reward was provided for every action the robot performed that guided it towards classifying fruit properly. This experiment will serve as the reference for the best possible performance in the task. We compared two other experiments to this one. An experiment where both top-down and bottom-up methods operated jointly and another experiment where the robot is only endowed with the classical top-down sub-goal learning mechanism. That is, the robot did not attempt to make use of latent knowledge contained in the perceptual classes it had created before. 

To simulate a robot that is learning different goals and perceptual classes at different points in its life, a learning curriculum was devised for the experiment. The idea is that the robot learns throughout its lifetime, achieveing different goals in different moments. This knowledge is not explicitly part of the final goal we will use for testing, but part of it may be reused. The following curreiculum phases we used:

\begin{enumerate}
    \item Exploratory phase where all task-related drives are satisfied. The weighing scale is turned off.
    \item The drive $D_{place}$ is satisfied when a fruit is placed in the center of the table. 
    \item Exploratory phase where all task-related drives are satisfied. The weighing scale is on, therefore it is active when fruits are placed on top.
    \item The drive $D_{classify}$ is satisfied when a fruit is properly classified. 
\end{enumerate}

Additionally, a cognitive drive based on effectance \cite{romero2020motivation} was provided to the robot by the designer so that when unexpected effects happen in the environment, a goal to recreate that effect is created for learning purposes. This goal will remain active until it is learned ($C > 0.95$), afterwards the robot will not try to replicate the effect again. The effects considered for this experiment are the activation of the binary sensors of the robot. By providing this cognitive drive, general goals are created, so that the robot leverages them in later phases of its lifetime.

\subsection{Results and Discussion}

The three experiments described were executed using a discrete event simulator in order to speed-up the data collection process. Each experiment was executed until the final goal for 750 trials of achieving the final goal. An upper limit of 20 iterations per trial was set, after which the world is reset. A total of 10 runs per experiment were performed. The first stage of the experiment was active for the first 20 trials, the second until trial 125, and the third phase started after 150 trials. 

\begin{figure}[ht]
  \centering
  \includegraphics[width=1.0\columnwidth]{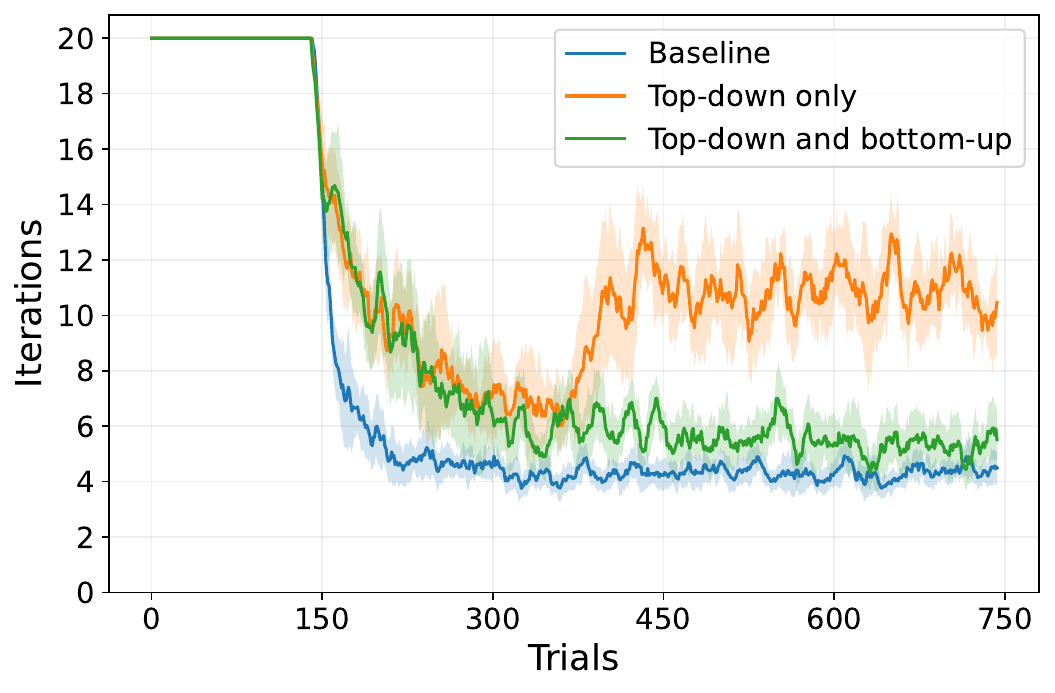}
  \caption{Performance comparison of the different approaches of sub-goal generation.}
  \label{fig:performance_comparison}
\end{figure}

The performance comparison of the different approaches is shown in Fig. \ref{fig:performance_comparison}. During the first three phases of the experiment, the final goal is not achieved and therefore every trial takes the maximum number of iterations. The top-down only experiment stabilizes at around 11 iterations per trial. The two-pronged approach stabilizes at approximately 5 iterations per trial, a performance similar to the baseline experiment which takes 4 iterations per trial on average. Bear in mind that in the reference experiment the robot is being given a reward every action it takes and in the other two only when the goal is reached. 

\begin{figure}[ht]
  \centering
  \includegraphics[width=0.9\columnwidth]{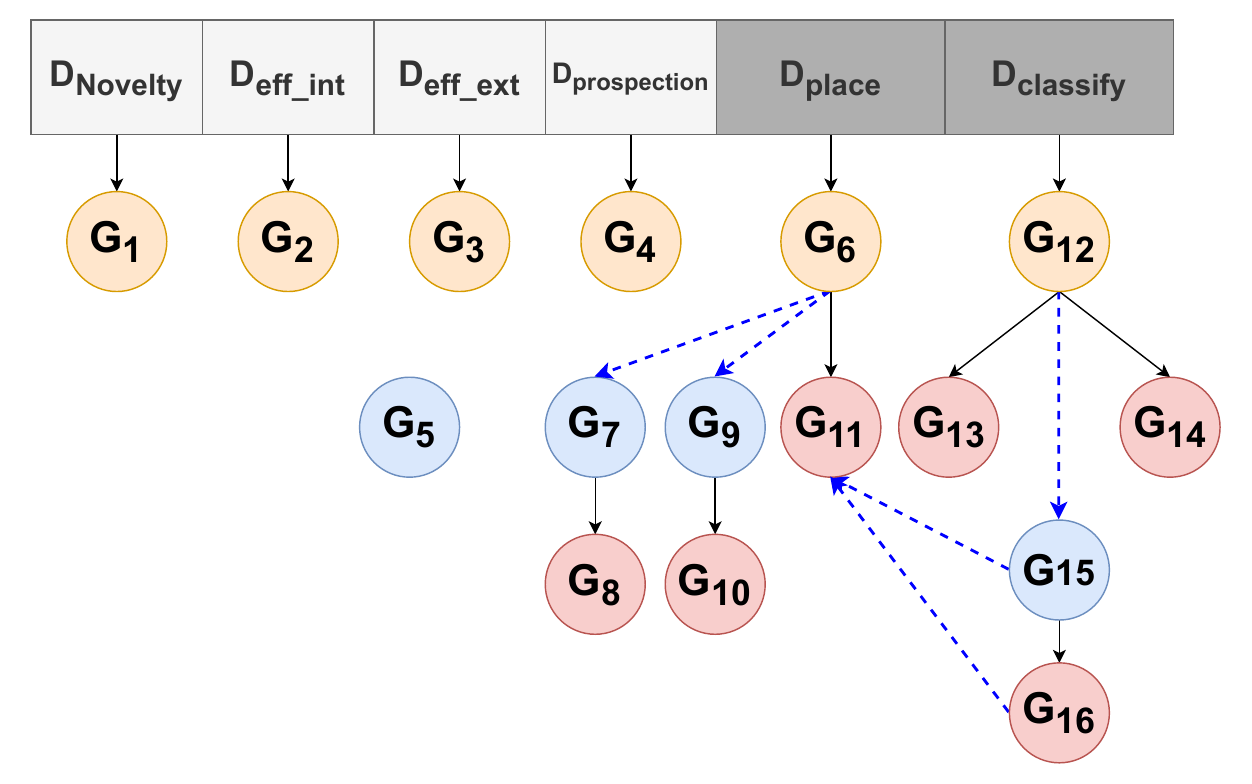}
  \caption{Goal tree after the execution of the Top-down + Bottom-up approach experiment. Goals represented in light blue were created by effects in the environment. Goals represented in red were created by the top-down sub-goal creation mechanism. Arrows in blue represent latent links found by the bottom-up prospection process.}
  \label{fig:goal_tree_full_exp}
\end{figure}

\begin{figure}[ht]
  \centering
  \includegraphics[width=0.8\columnwidth]{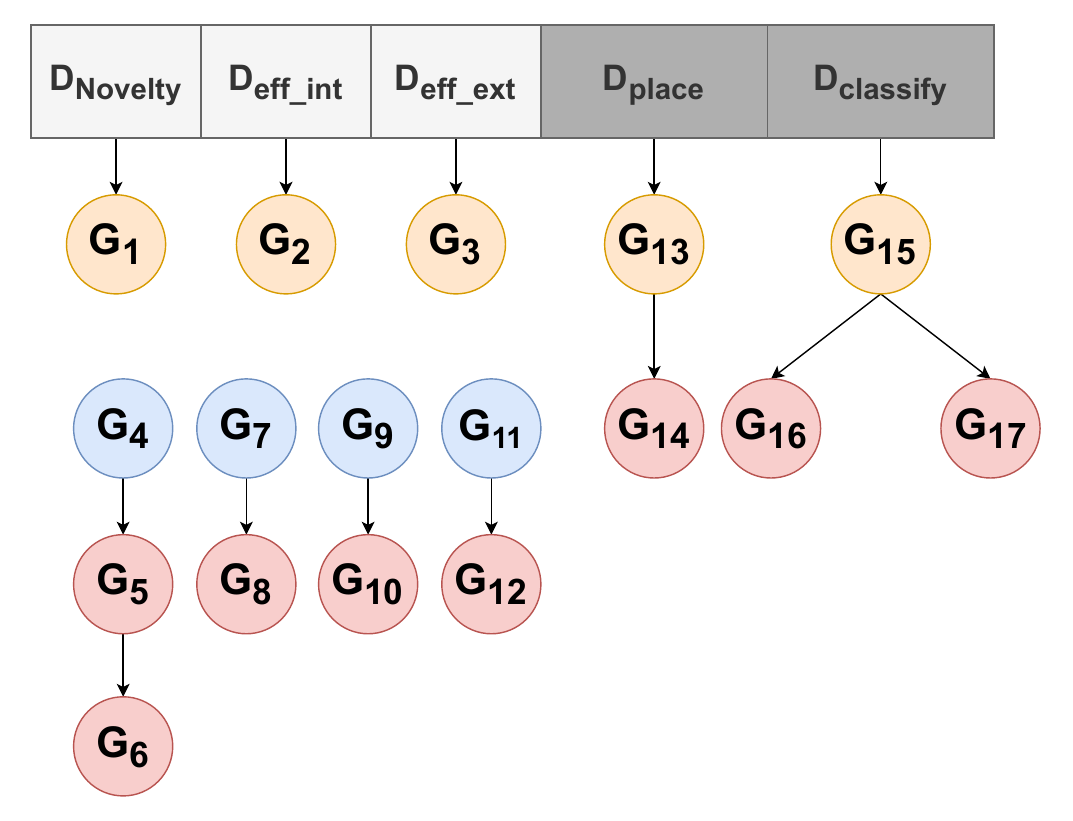}
  \caption{Goal tree after the execution of the top-down only experiment. Goals represented in light blue were created by effects in the environment. Goals represented in red were created by the top-down sub-goal creation mechanism.}
  \label{fig:goal_tree_topdown_exp}
\end{figure}

The final goal graph generated by the Top-down + Bottom-up experiment and the Top-down only experiment are shown in Fig. \ref{fig:goal_tree_full_exp} and Fig. \ref{fig:goal_tree_topdown_exp} respectively. Additionally, the interpretation of each of the goals of the graphs is presented in tables \ref{tab:goals_full_exp} and \ref{tab:goals_topdown_exp}. Note that the description given of the goals is derived from analyzing the goal tree and the spaces generated. This shows how by having a discrete representation for each of the goals in memory it is possible to interpret the knowledge obtained and possibly explain the decision process of the robot.

\begin{table}[ht]
\centering
\caption{Goals in the LTM after the execution of the Top-down + Bottom-up approach experiment.}
\label{tab:goals_full_exp}
\resizebox{0.85\columnwidth}{!}{%
\begin{tabular}{cl}
\hline
\multicolumn{1}{l}{\textbf{Goal}} & \textbf{Description}                        \\ \hline
$G_1$                             & Novel state achieved                        \\
$G_2$                             & Sub-goal created for learned P-Node         \\
$G_3$                             & Goal linked to environment effect created   \\
$G_4$                             & Sub-goal link created                       \\
$G_5$                             & Button light turned on                      \\
$G_6$                             & Fruit placed on table                       \\
$G_7$                             & Fruit grasped with left gripper             \\
$G_8$                             & Fruit grasped with right gripper            \\
$G_9$                             & Fruit grasped with right gripper            \\
$G_{10}$                            & Fruit grasped with left gripper             \\
$G_{11}$                            & Fruit grasped on any gripper                \\
$G_{12}$                            & Fruit classified properly                   \\
$G_{13}$                            & Good fruit tested                           \\
$G_{14}$                            & Bad fruit tested                            \\
$G_{15}$                            & Weighing scale activated                    \\
$G_{16}$                            & Fruit grasped on the same side as the scale \\ \hline
\end{tabular}%
}
\end{table}

\begin{table}[ht]
\centering
\caption{Goals in the LTM after the execution of the top-down only experiment.}
\label{tab:goals_topdown_exp}
\resizebox{0.85\columnwidth}{!}{%
\begin{tabular}{cl}
\hline
\multicolumn{1}{l}{\textbf{Goal}} & \textbf{Description}                        \\ \hline
$G_1$                             & Novel state achieved                        \\
$G_2$                             & Sub-goal created for learned P-Node         \\
$G_3$                             & Goal linked to environment effect created   \\
$G_4$                             & Button light turned on                      \\
$G_5$                             & Button light turned off                     \\
$G_6$                             & Button light turned on                      \\
$G_7$                             & Fruit grasped with left gripper             \\
$G_8$                             & Fruit grasped with right gripper            \\
$G_9$                             & Fruit grasped with right gripper            \\
$G_{10}$                            & Fruit grasped with left gripper             \\
$G_{11}$                            & Weighing scale activated                    \\
$G_{12}$                            & Fruit grasped on the same side as the scale \\
$G_{13}$                            & Fruit placed on table                       \\
$G_{14}$                            & Fruit grasped on any gripper                \\
$G_{15}$                            & Fruit classified properly                   \\
$G_{16}$                            & Good fruit tested                           \\
$G_{17}$                            & Bad fruit tested                            \\ \hline
\end{tabular}%
}
\end{table}

Both the performance graph and goal graph generated shows how the proposed method allows the robot to leverage the goals learned in the first phases of the experiment. For example, when pursuing the goal of placing a fruit on the table ($G_6$, table \ref{tab:goals_full_exp}) the experience obtained from the effect of grasping objects with either of the grippers is used, as shown by the connections to $G_7$ and $G_9$. Using this knowledge still allows to create a representation of the general sub-goal that leads to placing a fruit on the table, which is having a piece of fruit grasped with any gripper. This general sub-goal is then used as a steppingstone when learning to activate the scale, as shown by the connection of $G_{15}$ to $G_{11}$.

When the experiment is in the classifying stage, pursuing the goal of activating the scale bootstraps the learning of the classification P-Nodes. This is because the robot is more likely to encounter situations where it has to accept or discard a fruit rather than if all the previous actions were performed by exploration. An important remark is that the learning curriculum was designed such that the last phase starts when the learning process of $G_{15}$ (top-down + bottom-up) or $G_{11}$ (top-down only) goals is still not finished, therefore the robot is still motivated to pursue that goal. This is specially notorious in the performance graph of the top-down only experiment that shows decreased performance around trial 375, which is determined by the moment where $G_{11}$ gets learned and is no longer active. Fig. \ref{fig:goal_tree_topdown_exp} shows how the possible sub-goals remain in memory but are not linked to any other goal. This showcases the importance of the bottom-up mechanism where the sub-goal gets activation directly from the final goal after being linked, and therefore, the bootstrapping effects becomes permanent. Furthermore, the performance in the top-down only experiment does not improve after the environment effect goals (represented in light blue in Fig. \ref{fig:goal_tree_full_exp} and Fig. \ref{fig:goal_tree_topdown_exp}) are learned. This is because the top-down mechanism does not generate sub-goals beyond $G_{16}$ and $G_{17}$, as the P-Nodes related to these goals never reach the confidence threshold.  
 
\begin{figure}[ht]
  \centering
  \includegraphics[width=0.9\columnwidth]{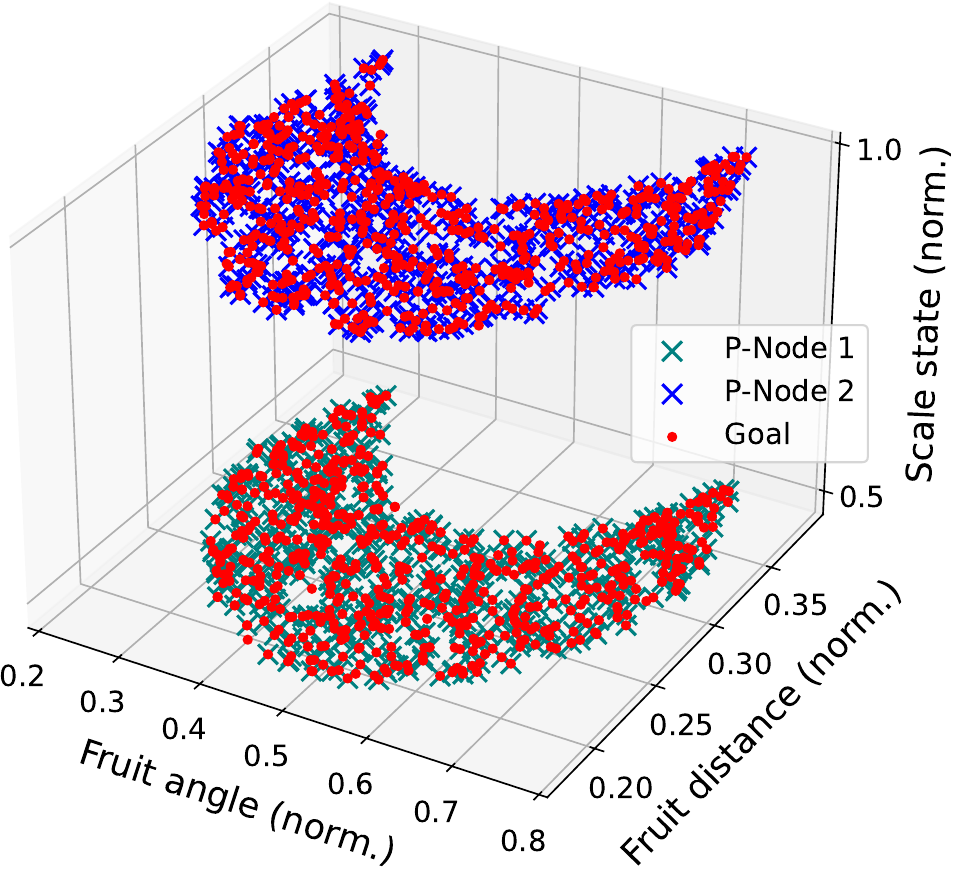}
  \caption{Comparison between the activation areas of the goal \textit{weighing scale activated} and the P-Nodes \textit{bad fruit tested} (P-Node 1) and \textit{good fruit tested} (P-Node 2).}
  \label{fig:space_comparison}
\end{figure}

Fig. \ref{fig:space_comparison} shows the activation areas of the goal related to placing fruit on the weighing scale ($G_{15}$, table \ref{tab:goals_full_exp}) and the P-Nodes related to the \textit{accept\_fruit} and \textit{discard\_fruit} policies. In the figure it can be seen how there are two planes which correspond to the possible positions of the scale and the two levels of fruit state (good=0.5, bad=1.0) the goal covers both areas as it does not discriminate between good or bad fruits. On the other hand, the P-Nodes are dependent on the state of the fruit and therefore their activation areas lie on one of the planes each. This correspondence is detected by the bottom-up mechanism and triggers the link between $G_{12}$ and $G_{15}$.

\section{Conclusions}
\label{sec:conclusion}
This paper addresses the problem of sub-goal generation when robots need to achieve final goals that require a sequence of steps in lifelong open-ended settings, which may be complex. To this end, we show how leveraging the potential of more traditional top-down sub-goal discovery approaches that are based on exploring paths towards the starting state of a goal in order to convert this state into a sub-goal and progressively construct a hierarchy of sub-goals is not enough. As the sub-goal chain grows, this process becomes more difficult and inefficient, especially when the robot has had previous experience in other domains or tasks. 

To address this issue, we propose a second simultaneous bottom-up approach that seeks to explore the knowledge on perceptual classes the robot already has, to find latent relationships between perceptual classes (including goals). These can then be used to determine sub-goal paths towards a goal in a more efficient manner.

We have tested the two-pronged approach we propose and compared it to a traditional top-down approach and a baseline case in which the robot was guided by rewards towards the final goal. The results were quite encouraging. Our proposal achieved an efficiency that was very close to the optimal one in the reference case but only rewarding the final goal. In fact, in terms of iterations needed to reach the final goal, the final result is two and a half times better than the traditional top-down approach by itself. 

Obviously these are just the first steps in this research. There is still a need to develop more powerful and efficient algorithms for the discovery of these latent relationships within the perceptual class memories as well as leverage other intrinsic motivations to facilitate the process so that more complex cases can be addressed.



\bibliographystyle{unsrt}  
\bibliography{IJCNN2025}  






\end{document}